\documentclass[letterpaper, 10 pt, conference]{ieeeconf}  

\IEEEoverridecommandlockouts                              

\usepackage{xcolor}

\usepackage{amsmath} 
\usepackage{amssymb}  
\usepackage{mathtools}
\usepackage{kotex}
\usepackage{multicol}
\usepackage{multirow}
\usepackage{booktabs}
\usepackage{pifont}
\usepackage{color,soul}

\usepackage{graphicx}
\usepackage{hhline}
\usepackage{dsfont}
\usepackage{bbm}
\usepackage{rotating}
\usepackage{makecell}
\usepackage{cite}
\usepackage[table]{xcolor}
\usepackage[dvipsnames]{xcolor}

\newcommand{\cmark}{\ding{51}}

\title{\LARGE \bf
MapTCL: Temporal Consistency Learning via Bidirectional Alignment for Vectorized HD Map Construction
}

\author{
Hyeonseo Kim$^{1}$,
Juyeb Shin$^{1}$,
Hyeonjun Jeong$^{2}$,
Hiwon Shin$^{1}$,
and Dongsuk Kum$^{2}$%
\thanks{This work was supported by Advanced GPU Utilization Support Program and the National Research Foundation of Korea (NRF) grant funded by the Korea government (MSIT) (RS-2026-25479609).}
\thanks{$^{1}$ H. Kim, J. Shin and H. Shin are with the Robotics Program, Korea Advanced Institute of Science and Technology (KAIST), Daejeon 34141, South Korea (email: {\tt\small hyeonseo.kim@kaist.ac.kr}, {\tt\small juyebshin@kaist.ac.kr}, {\tt\small hiwon.shin@kaist.ac.kr})}
\thanks{$^{2}$H. Jeong and D. Kum are with the Cho Chun Shik Graduate School of Mobility, Korea Advanced Institute of Science and Technology (KAIST), Daejeon 34141, South Korea (email: {\tt\small hyeonjun.jeong@kaist.ac.kr}, {\tt\small dskum@kaist.ac.kr})}
}

\begin{document}

\maketitle
\thispagestyle{empty}
\pagestyle{empty}

\begin{abstract}
Constructing reliable online HD maps remains challenging in dynamic urban environments due to moving objects and occlusions. 
While recent works employ feature-level temporal fusion to address this, they rely solely on per-frame ground truth supervision.
Consequently, they lack an explicit objective to directly penalize the geometric noise and temporal jitter between consecutive online HD maps.
To address this, we propose MapTCL, an auxiliary training strategy that formulates temporal consistency loss between current and past frames via bidirectional alignment.
Specifically, Bidirectional Vector Consistency Learning (BVCL) models the geometric and semantic discrepancies between associated past and current vector instances as an auxiliary loss.
We also employ Raster map Consistency Learning (RCL) as an additional loss to stabilize dense BEV features.
By jointly training with these dual losses, MapTCL improves the temporal stability of generated HD maps.
Extensive experiments on two standard benchmarks demonstrate the effectiveness of our approach. As a versatile plug-and-play module, MapTCL consistently enhances existing baseline models, achieving gains of +3.7 mAP \& +2.8 C-mAP on nuScenes and +3.1 mAP \& +2.5 C-mAP on Argoverse 2 without additional inference overhead.

\end{abstract}

\section{INTRODUCTION}

\label{sec:intro}



%

High-Definition (HD) maps provide centimeter-level accuracy and vectorized semantic representations of static objects, which are crucial for reliable localization\cite{jeong2024multi, chalvatzaras2022survey} and path planning\cite{casas2021mp3, jiang2023vad} in autonomous driving.
Traditionally, HD maps have been generated through offline SLAM-based methods~\cite{zhang2014loam, shan2018lego} with a mobile mapping system, which involves complex pipelines and manual annotation. This process also requires human verification and regular updates, leading to high maintenance costs and labor. 
To mitigate these issues, online HD map construction~\cite{li2022hdmapnet,liu2023vectormapnet,liao2022maptr,shin2025instagram,liao2024maptrv2,choi2024mask2map,yuan2024streammapnet,wang2024stream,chen2024maptracker,kim2025unveiling} 
using onboard sensors has emerged as a promising alternative. 
This approach leverages neural networks with onboard sensors to detect map elements and construct vectorized local HD maps without manual annotation. 



\begin{figure}[t]
\centering
\vspace{2pt}
\includegraphics[scale=0.50]{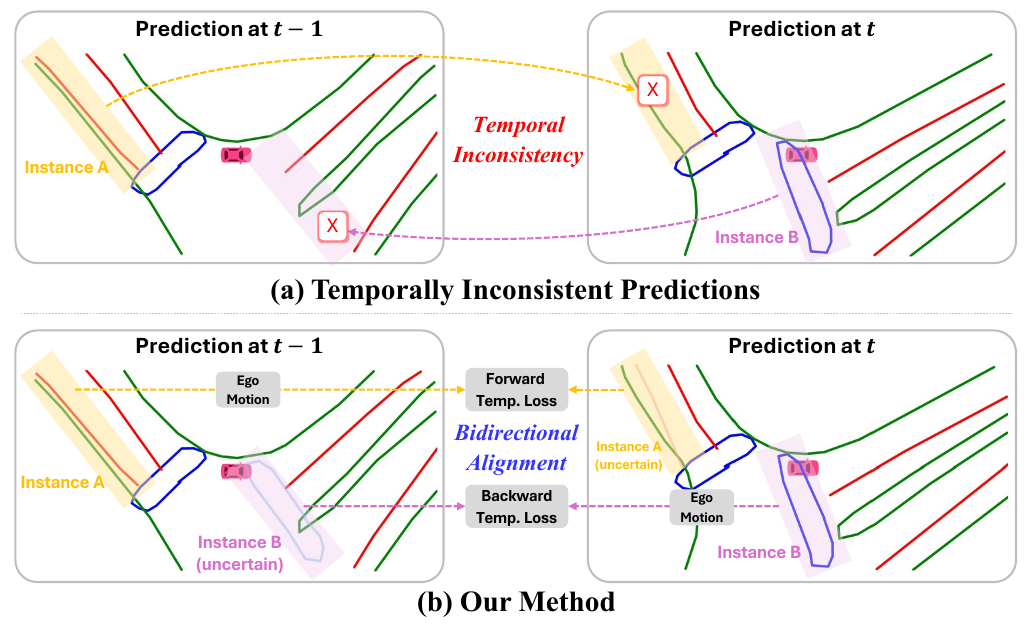}
\vspace{-20 pt}
\caption{Motivation and core concept of MapTCL.
(a) Examples of temporally inconsistent predictions in online vectorized HD maps. The yellow and purple boxes highlight vector instances that fail to maintain continuity across consecutive frames. 
(b) The core concept of MapTCL. 
The bidirectional alignment loss enhances temporal consistency.
}
\label{figure:Intro}
\vspace{-20pt}
\end{figure}

However, early works~\cite{li2022hdmapnet,liu2023vectormapnet,shin2025instagram,liao2022maptr,liao2024maptrv2} predict each local map independently, without explicitly modeling temporal consistency.
As shown in Fig.~\ref{figure:Intro} (a), this often leads to unstable vector predictions across consecutive frames, such as spatial jitter, flickering instances, or temporary disappearance of map elements, especially under dynamic occlusions and long-range perception.
To mitigate this, various approaches~\cite{yuan2024streammapnet,kim2025unveiling,chen2024maptracker} have been proposed to utilize temporal information, such as aggregating past BEV features and queries, interacting with clip-level memory tokens and tracking historical queries.
Despite these architectural advancements, they mainly optimize each predicted HD map against the ground truth of the same timestamp, without directly comparing predictions across time. 
Therefore, even if the ground-truth annotations are temporally consistent, the model is not explicitly penalized when the same physical map element is predicted with different geometric and semantic continuity in sequential frames.
This makes the output vulnerable to spatial jitter, flickering, and unstable vector instances.

To address these limitations, we propose MapTCL, an auxiliary training strategy that reinforces temporal consistency and reduces frame-to-frame noise in local HD maps. 
MapTCL applies a dual temporal constraint across two prediction-level representations: Bidirectional Vector Consistency Learning (BVCL) for vector elements and Raster Consistency Learning (RCL) for  raster maps.
At its core, BVCL directly penalizes the geometric and semantic discrepancies between the model's past and current vector predictions, as demonstrated in Fig.~\ref{figure:Intro} (b). To formulate this accurately without propagating historical errors, we introduce Bidirectional Temporal Matching (BTM).
BTM first filters only high-confidence vector instances from consecutive vectorized maps. These selected vectors are then transformed into each other's coordinate system and matched bidirectionally. Finally, the geometric and semantic differences between the matched elements are formulated as an auxiliary loss. 
MapTCL also integrates RCL as a supplementary training to stabilize the BEV feature consistency. In this module, BEV features from the image backbone are processed by a segmentation head to generate dense raster maps. The additional auxiliary loss is then defined as the discrepancy between the current and past raster maps.
By jointly optimizing these two auxiliary losses, MapTCL enforces self-consistency across long-term frames and effectively suppresses temporal jitter of online HD map predictions.
Furthermore, existing temporal fusion methods~\cite{yuan2024streammapnet, chen2024maptracker,kim2025unveiling} increase computational and memory costs during inference, thereby degrading the real-time performance required for online HD map construction.
In contrast, MapTCL is a plug-and-play training framework that introduces no additional inference overhead.
We conduct extensive experiments by applying MapTCL to various baseline and SOTA models on the nuScenes~\cite{caesar2020nuscenes} and Argoverse2~\cite{wilson2023argoverse} datasets. Across multiple metrics, our method consistently improves both accuracy and temporal consistency without additional inference costs.
In summary, our contributions are as follows:
\begin{itemize}
    \item 
    Bidirectional Vector Consistency Learning (BVCL) is proposed as a novel mechanism to enforce temporal consistency by formulating bidirectional alignment between vector instances across consecutive HD maps.
    
\vspace{2pt}
    \item 
    MapTCL is introduced as a plug-and-play auxiliary training framework. By jointly applying BVCL and RCL to baseline models, explicit supervision is provided to ensure robust and temporally coherent HD map construction.
    
\vspace{2pt}
    \item Our method achieves improved performance on two standard benchmarks (nuScenes and Argoverse2) with various baseline models, without introducing additional computational and memory overhead during inference.
\end{itemize}

\section{RELATED WORK}
\label{sec:related_works}


\subsection{Online HD Map Construction}
\vspace{-2pt}
\indent With the rapid advancements in deep learning, online HD map construction using onboard sensors has gained significant attention as a cost-effective alternative to traditional HD map construction. HDMapNet~\cite{li2022hdmapnet} generates a rasterized map through BEV segmentation and converts it into a vector representation via post-processing. However, this process is heuristic and incurs additional processing.
VectorMapNet~\cite{liu2023vectormapnet} introduces the first end-to-end learning framework for online HD mapping, employing a coarse-to-fine two-stage approach with DETR~\cite{carion2020end} network. MapTR~\cite{liao2022maptr} improves the VectorMapNet pipeline by introducing hierarchical transformer queries and permutation-equivalent modeling into a DETR-based one-stage architecture.
MapTRv2~\cite{liao2024maptrv2} further improves MapTR by modifying the decoder architecture and incorporating multiple auxiliary training strategies, which significantly enhances HD map detection performance. 
InstaGraM~\cite{shin2025instagram} proposes a graph-based end-to-end network for efficient HD mapping. 
Mask2Map~\cite{choi2024mask2map} employs BEV segmentation masks and mask-aware queries to extract geometric features and positional context.
MapQR~\cite{liu2024mapqr} improves spatial accuracy by enhancing instance-level queries through a scatter-and-gather query design.
However, since these methods do not exploit historical context, they remain vulnerable to temporal inconsistency in complex scenarios, such as dynamic occlusions or long-range perception.

\vspace{-4pt}
\subsection{Temporal Learning for HD Map Construction}
\vspace{-2pt}

Utilizing temporal information is essential for constructing robust and temporally consistent HD maps. 
StreamMapNet~\cite{yuan2024streammapnet} pioneers the integration of temporal modeling into online HD map construction by streaming historical latent through GRU (Gated Recurrent Unit)-based BEV feature fusion and streaming queries.
MapTracker~\cite{chen2024maptracker}, inspired by MOTR~\cite{zeng2022motr}, formulates HD mapping as a tracking mechanism by generating tracked ground truth and performing bipartite matching based on these tracked annotations. 
It also stores long-term past BEV features and vector elements in a memory buffer and strengthens temporal continuity by fusing them.
MapUnveiler~\cite{kim2025unveiling} utilizes long-term clip-level tokens to enhance HD map consistency. Its Inter-clip Unveiler module stores key token features extracted from previous clips and integrates them with the current clip’s features via an attention mechanism. 
Despite improved performance, these methods fundamentally rely on per-frame ground truth supervision. 
Furthermore, aggregating a large amount of past features leads to high memory usage and increased computational costs during inference.
However, as an auxiliary training method, MapTCL strengthens temporal consistency without extra computational and memory cost during inference.
SQD-MapNet~\cite{wang2024stream}, inspired by DN-DETR~\cite{li2022dn}, strengthens the temporal consistency of StreamMapNet by denoising ground-truth queries. 
DTCLMapper~\cite{li2024dtclmapper} enhances temporal instance consistency via contrastive learning and further enforces geometric consistency across frames using a Map Occupancy loss on ego-aligned rasterized maps.
In contrast, our method enforces temporal consistency via bidirectional matching of vector elements, directly minimizing geometric and semantic inconsistencies between past and current predictions, and complementing it with raster-level consistency learning.

\begin{figure*}[t]
\begin{center} 
\includegraphics[scale=0.90]{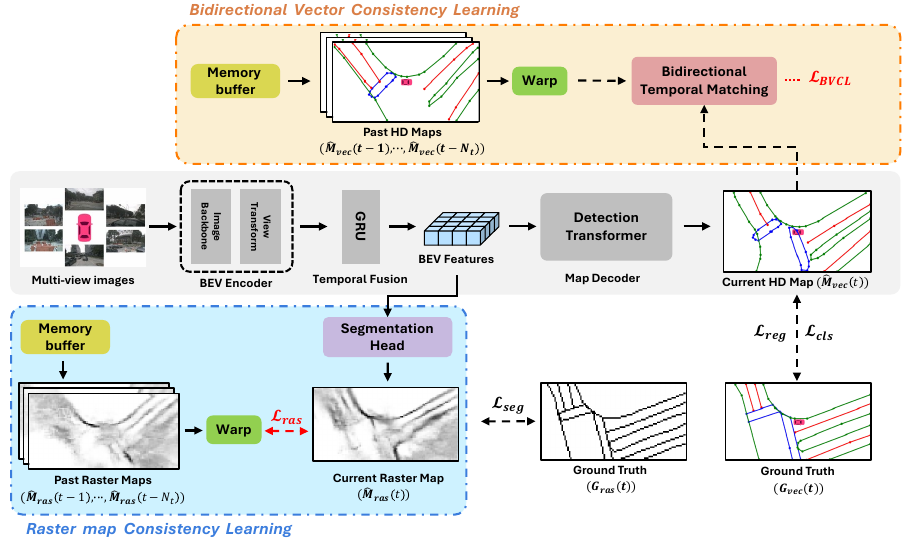}
\vspace{-10pt}
\caption{
Overall pipeline of MapTCL. The pipeline takes multi-view images as input, encodes them into BEV features, and employs a map decoder to regress vectorized HD maps. The vector and raster maps are stored in the memory buffer at each iteration and subsequently utilized in both BVCL and RCL. These processes are removed during inference.
}
\label{figure:main}
\vspace{-20pt}
\end{center}
\end{figure*}


\section{METHODOLOGY}
\label{sec:method}

\subsection{Overview}
MapTCL aims to model and enforce temporal consistency between sequential frames through an auxiliary training strategy. In this paper, we adopt StreamMapNet~\cite{yuan2024streammapnet} as our baseline for BEV-based online HD map construction.
Fig.~\ref{figure:main} shows the typical BEV-based online HD map construction pipeline, which utilizes synchronized multi-view images~\cite{caesar2020nuscenes,wilson2023argoverse}, BEV (Bird's Eye View) encoder~\cite{li2024bevformer} for view transformation and a map decoder~\cite{carion2020end} for vector regression. During each training iteration, the memory buffer stores and updates past HD maps $\boldsymbol{\hat{M}_{vec}}(t-1),..., \boldsymbol{\hat{M}_{vec}}(t-N_t)$ and raster maps $\boldsymbol{\hat{M}_{ras}}(t-1),...,\boldsymbol{\hat{M}_{ras}}(t-N_t)$ generated from ${N_t}$ historical frames, where $t$ denotes the current time. These memories are then used for temporal consistency learning, which includes Bidirectional Vector Consistency Learning (BVCL) and Raster map Consistency Learning (RCL). These processes are removed during inference without incurring any additional computational cost and memory.
We first explain BEV-based online HD map construction in section~\ref{baseline_section}, followed by the introduction of our method into this framework: (1) Bidirectional Vector Consistency Learning (BVCL) in section~\ref{BVCL}, and (2) Raster map Consistency Learning (RCL) in section~\ref{RCL}.

\vspace{-4pt}
\subsection{BEV-based Online HD Map Construction}
\label{baseline_section}

The BEV encoder, consisting of an image backbone and a view transformation network~\cite{li2024bevformer}, extracts BEV features from multi-view images. These features are fused with the past one stored in a memory buffer using GRU (Gated Recurrent Unit).
These are then fed into a DETR~\cite{carion2020end}-based map decoder, where learnable queries attend to the features to predict vectorized map elements. Through bipartite matching, predicted $N$ instances are matched with the ground truth $\boldsymbol{G_{vec}}(t)$ at time $t$.
The predicted HD map is represented as $N$ vectorized instances $\boldsymbol{\hat{M}_{vec}}(t)=\left\{\left(\hat{v}_i\right)\right\}_{i=0}^{N-1}$ and each map instance $\hat{v}_i=\left(\boldsymbol{\hat{C}_i}, \boldsymbol{\hat{P}_i}\right)$ consists of class distribution (pedestrian crossing, lane divider, road boundary) and position of $N_p$ vector points, where $\boldsymbol{\hat{C}}=\left\{\hat{c}_j\right\}_{j=1}^{3}$ and $\boldsymbol{\hat{P}}=\left\{\left(\hat{x}_j, \hat{y}_j \right)\right\}_{j=0}^{N_p-1}$.

\vspace{-3pt}
\subsection{Bidirectional Vector Consistency Learning}
\label{BVCL}
\begin{figure}[t]
\begin{center}
\includegraphics[scale=0.47]{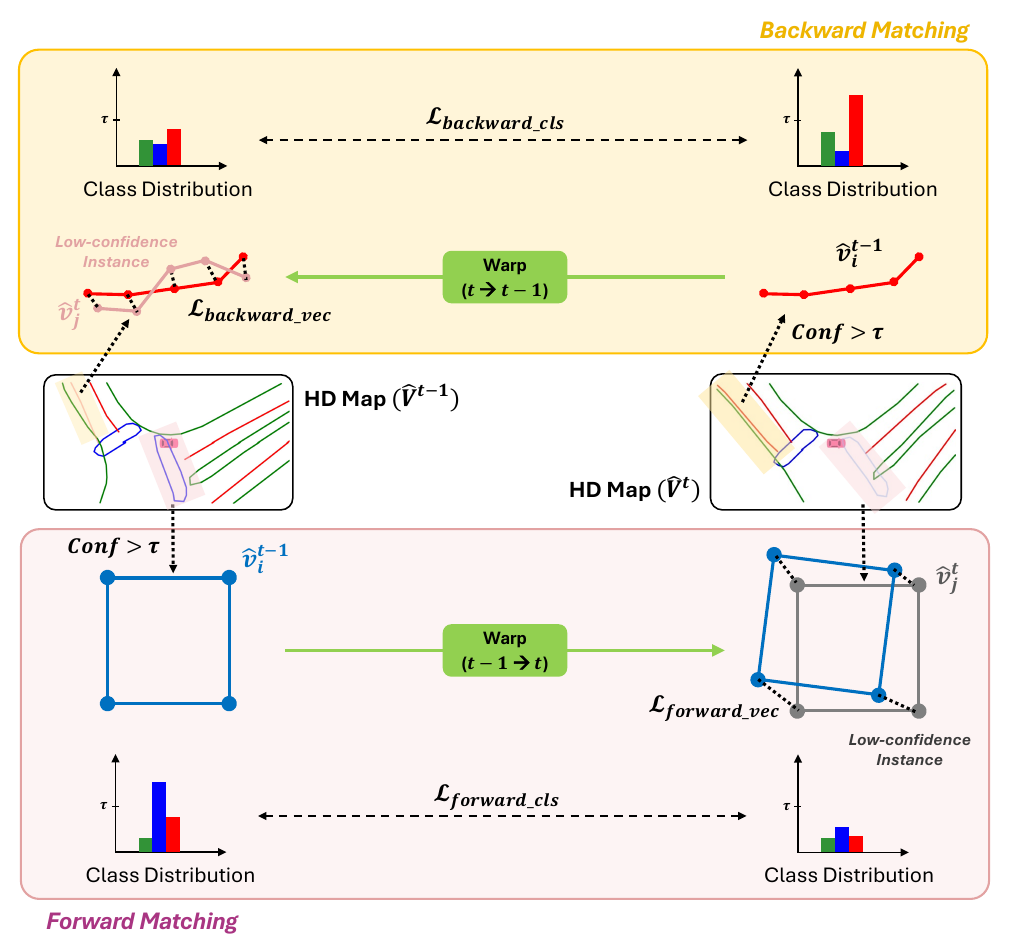}
\vspace{-10pt}
\caption{Bidirectional Temporal Matching mechanism. High-confidence elements from past vectorized HD maps are warped to the current frame using ego pose and compared with current predictions to compute distance and class distribution losses. This process is performed bidirectionally.
} 
\label{figure:BTM}
\vspace{-25pt}
\end{center}
\end{figure}

Vector instances in consecutive local HD maps should remain temporally consistent. To this end, we draw inspiration from cycle consistency in 3D scene flow estimation~\cite{mittal2020just}, which imposes a strict geometric constraint by enforcing that 3D points tracked forward and backward across frames return to their original coordinates. By adapting this cycle alignment principle to vector-based HD mapping, we propose Bidirectional Vector Consistency Learning (BVCL). 
In contrast to existing methods supervised only by per-frame ground truth, BVCL enforces forward–backward consistency by jointly aligning the geometric and semantic attributes of vector instances across frames.
To match misaligned vector instances from different timestamps, we introduce a novel Bidirectional Temporal Matching (BTM) method.
As illustrated in Fig. \ref{figure:BTM}, BTM mechanism matches confident vector elements from past and current frames in both directions to enhance temporal association, thereby mitigating the risk of error propagation across frames.
In the predicted past local HD map at time $t-N_t$, we select map elements with confidence scores above $\tau$ and transform them into the current coordinate system using the ego vehicle's motion. We then conduct bipartite matching between these warped reliable elements and all predicted $N$ instances in the current HD maps. 
This matching process is also applied in the opposite direction. We transform the current predicted instances into the coordinate system of each past frame $t-N_t$ using the inverse ego-motion.
Since the set of confident vector elements in past and current HD maps may differ due to temporal inconsistency, the matched elements in each direction may also vary.
By performing bidirectional matching and modeling the discrepancies as a loss, we reinforce consistent associations between consecutive vector instances while weakening inconsistent ones.
The matching process consists of instance-level and point-level matching, following the hierarchical matching strategy introduced in MapTR~\cite{liao2022maptr}. First, we determine the optimal instance-level assignment $\hat{\pi}$ between a set of current $N$ map elements $\{(\hat{v}_i^t)\}_{i=0}^{N-1}$ from $\boldsymbol{\hat{M}_{vec}}(t)$ and warped previous $K$ map element $\{(\hat{v}^{t-N_t}_{i})\}_{i=0}^{K-1}$ from $\boldsymbol{\hat{M}_{vec}}(t-N_t)$. $\hat{\pi} \in \Pi_K$ is an optimal permutation of $K$ map elements with the lowest matching cost of instance-level matching as defined in Eq. \ref{eq:ins-matching}. 
The pair-wise matching cost $\mathcal{L}_{\text{ins\_match}}$ is computed by considering both the class score and the positional distance of the vector instances.

\begin{equation}
\label{eq:ins-matching}
\vspace{-20pt}
\begin{aligned}
\hat{\pi} = \arg\min_{\pi \in \Pi_K} \sum_{i=1}^{K} \mathcal{L}_{\text{ins\_match}} \left(\hat{v}^{t-N_t}_{\pi(i)},\hat{v}^{t}_{i}\right),
\end{aligned}
\end{equation}

\vspace{5pt}
\begin{equation}
\label{eq:ins-matching2}
\begin{aligned}
\mathcal{L}_{\text {ins\_match }}\left(\hat{v}^{t-N_t}_{\pi(i)},\hat{v}^{t}_{i}\right) = & 
\mathcal{L}_{\text {cls }}\left(\boldsymbol{\hat{C}}^{t-N_t}_{\pi(i)}, \boldsymbol{\hat{C}}^{t}_i \right) \\
& +\mathcal{L}_{\text {pos}}\left(\boldsymbol{\hat{P}}^{t-N_t}_{\pi(i)}, \boldsymbol{\hat{P}}^{t}_i\right).
\end{aligned}
\end{equation}

\noindent 
We represent the class prediction of each instance as a multi-class probability and define the class matching cost $\mathcal{L}_{\text{cls}}$ as the L1 distance.
$\mathcal{L}_{\text{pos}}$ defined as SmoothL1 Loss that formulates positional distance between them. Hungarian algorithm~\cite{kuhn1955hungarian} is applied based on these costs to identify optimal instance assignment $\hat{\pi}$. 
After the instance-level matching, point-level matching is performed based on the assigned instance $\hat{v}^{t-N_t}_{\hat{\pi}(i)}$ to find the optimal point-to-point correspondence. The optimal assignment of the points $\hat{\gamma} \in \Gamma$ is selected from point-level matching cost:
\begin{equation}
\label{eq:pts_matching}
\begin{aligned}
\hat{\gamma}=\arg \min _{\gamma \in \Gamma} \sum_{j=1}^{N_p} \mathcal{L}_{\text{pos}}\left(\hat{p}^{t-N_t}_{j}, \hat{p}^{t}_{\gamma(j)}\right),
\end{aligned}
\end{equation}

\noindent where $\mathcal{L}_{\text{pos}}$ is the SmoothL1 Loss.
This forward matching step is then performed in reverse. The details of the loss functions are provided in section~\ref{Loss}.

\vspace{-5pt}
\subsection{Raster map Consistency Learning}
\label{RCL}
\vspace{-2pt}
BEV features extracted by the BEV encoder provide a structured and spatially unified representation of the surrounding environment, which is crucial for regressing vectorized instances. To promote temporal coherence in these representations, we introduce Raster map Consistency Learning (RCL).
Inspired by temporal consistency losses that penalize pixel-level discrepancies across frames in video segmentation~\cite{varghese2021unsupervised} and matting~\cite{sun2021matting}, RCL adapts dense cross-frame alignment to the BEV space. 
Following the MapTRv2~\cite{liao2024maptrv2} approach, we first generate a binary raster map $\boldsymbol{\hat{M}_{ras}}(t)$ by segmentation head $\phi_{BEVSeg}$, as illustrated in Fig.~\ref{figure:main}.
The raster map is supervised with the rasterized ground truth map $\boldsymbol{G}_{\mathrm{ras}}(t)$, providing dense supervision that encourages the model to capture the global structure and geometry of the map.
Building on this, the RCL further enforces temporal consistency by leveraging the past raster maps $\boldsymbol{\hat{M}_{ras}}(t-N_t)$ from the memory buffer. We warp them to the current coordinate frame and compute the inconsistency between the current raster map $\boldsymbol{\hat{M}_{ras}}(t)$ and the historical one. This inconsistency is formulated as an additional auxiliary loss.
We leave the detailed loss function of RCL to section~\ref{Loss}. 

\vspace{-5pt}
\subsection{Training Loss}
\label{Loss}
MapTCL adopts an end-to-end training based on map loss with GT supervision and our auxiliary loss that comprise BVCL and RCL.

\textbf{Map Loss.} Map loss is weighted sum of several basic losses from baseline method (StreamMapNet~\cite{yuan2024streammapnet}), which are regression loss, classification loss and transformation loss:
\begin{equation}
\label{eq:basic loss}
\begin{aligned}
\mathcal{L}_{\text{map}} = \lambda_1 \mathcal{L}_{\text{reg}} + \lambda_2 \mathcal{L}_{\text{cls}} + \lambda_3 \mathcal{L}_{\text{trans}},
\end{aligned}
\end{equation}
\noindent where $\lambda_1$, $\lambda_2$ and $\lambda_3$ are the hyper-parameters. The regression loss and transformation loss are computed using SmoothL1 Loss, while the classification loss is calculated using Focal Loss.

\vspace{2pt}
\textbf{BVCL Loss.} BVCL follows both the forward and backward processes, as described in Sec.~\ref{BVCL}. At each stage, the point-to-point loss ($\mathcal{L}_{\text{forward\_vec}}$ and $\mathcal{L}_{\text{backward\_vec}}$) and class distribution loss ($\mathcal{L}_{\text{forward\_cls}}$ and $\mathcal{L}_{\text{backward\_cls}}$) between matched instances are defined as follows:
\begin{align}
\mathcal{L}_{\text{BVCL}} &= \mathcal{L}_{\text{forward\_vec}} + \mathcal{L}_{\text{forward\_cls}} \notag \\
&\quad + \mathcal{L}_{\text{backward\_vec}} + \mathcal{L}_{\text{backward\_cls}}.
\label{eq:BVCL loss}
\end{align}
We employ BTM with Hungarian algorithm~\cite{kuhn1955hungarian} to find an optimal instance-level and point-level assignment ($\hat{\pi}$ and $\hat{\gamma}$) between current and past $N_t$ frames. 
$\mathcal{L}_{\text{forward\_vec}}$ and $\mathcal{L}_{\text{forward\_cls}}$ are formulated as follows:
\begin{equation}
\begin{aligned}
    \mathcal{L}_{\text{forward\_vec}}
    = \sum_{n=1}^{N_t} w_n \sum_{i=1}^{K} \sum_{j=1}^{N_p}
    \mathcal{L}_{\text{pos}}\!\left(
    {\hat{p}}^{t-n}_{\hat{\pi}(i),\hat{\gamma}_i(j)},\,
    \hat{p}^{t}_{i,j}
    \right),
\end{aligned}
\end{equation}
\begin{equation}
\begin{aligned}
    \mathcal{L}_{\text{forward\_cls}} 
    = \sum_{n=1}^{N_t} u_n \sum_{i=1}^{K} 
    \mathcal{L}_{\text{KL}}\left(\hat{\textbf{C}}^{t-n}_{\hat{\pi}(i)}, \hat{\textbf{C}}^{t}_i\right),
\end{aligned}
\end{equation}
where the point-to-point loss uses SmoothL1 Loss, and the class distribution loss is defined by the KL-divergence. The $w_n$ and $u_n$ denote the temporal weights for positional and semantic consistency at the $t-N_t$ frame, respectively. We assign higher weight to temporally closer frames for more reliable alignment. The $\mathcal{L}_{\text{backward\_vec}}$ and $\mathcal{L}_{\text{backward\_cls}}$ are computed by warping the current HD map instance to the current frame, using the same mechanism and weights as the forward process. 


\textbf{RCL Loss.}
In RCL, we introduce a BEV segmentation head, $\phi_\mathrm{BEVSeg}$, based on BEV features to generate a binary raster map. The segmentation loss is defined as a mask Focal loss~\cite{lin2017focal}, a common pixel loss in vector HD mapping approaches~\cite{liao2024maptrv2,chen2024maptracker}, computed between the predicted binary raster map and the rasterized ground truth:
\begin{equation}
\begin{aligned}
\mathcal{L}_{\mathrm{seg}}=\mathcal{L}_{\mathrm{Focal}}\left(\boldsymbol{\hat{M}}_{ras}(t), \boldsymbol{G}_{ras}(t)\right),
\end{aligned}
\end{equation}
\begin{equation}
\begin{aligned}
\boldsymbol{\hat{M}}_{ras}(t)=\phi_\mathrm{BEVSeg}\left(\mathcal{F}_{\mathrm{BEV}}\right).
\end{aligned}
\end{equation}
The predicted raster map is then compared with the raster maps from past frames to compute the inconsistency. The Raster map Consistency loss is defined as a mask Focal loss between current raster map and historical raster maps:
\begin{equation}
\begin{aligned}
\mathcal{L}_{\mathrm{ras}}=\sum_{n=1}^{N_t} \alpha_n\mathcal{L}_{\mathrm{Focal}}\left(\boldsymbol{\hat{M}}_{ras}(t-n), \boldsymbol{\hat{M}}_{ras}(t)\right).
\end{aligned}
\end{equation}
\noindent The weight $\alpha_n$ is assigned to emphasize temporally closer frames by giving them higher value. The total loss of RCL is as follows:
\begin{equation}
\label{eq:RCL loss}
\begin{aligned}
\mathcal{L}_{\text{RCL}} = \beta\mathcal{L}_{\text{seg}} + \mathcal{L}_{\text{ras}}.
\end{aligned}
\end{equation}

\textbf{Overall Loss.}
The overall loss is defined as the sum of the above losses:
\begin{equation}
\label{eq:overall loss}
\begin{aligned}
\mathcal{L} = \mathcal{L}_{\text{map}} + \mathcal{L}_{\text{BVCL}} + \mathcal{L}_{\text{RCL}}.
\end{aligned}
\end{equation}


\section{Experiments}
\label{sec:experiments}

\begin{table*}[t]
\caption{Comparisons on nuScenes and Argoverse2 validation sets using non-overlapping data split within $60m\times30m$ range. FPS is measured using a single NVIDIA RTX 3090 GPU. $\dagger$ results reproduced in our environment using available code, and $^*$ results from the corresponding papers. - indicates unavailable results (C-mAP not reported and FPS omitted due to different evaluation environments).}
\vspace{-7pt}
\label{tab:3_Comparison_newsplit}
\begin{center}

    \resizebox{0.75\textwidth}{!}
    {
    \setlength{\tabcolsep}{6pt}
    \renewcommand{\arraystretch}{1.3}
    \begin{tabular}{l|l|cc|cccc|c|c}
        \toprule[1pt]
        Dataset & Method & Epochs & \makecell{Fused \\ Frames} & AP$_{\textit{ped}}$ & AP$_{\textit{div.}}$ & AP$_{\textit{bound.}}$ & mAP & C-mAP & FPS  \\
        \hline
        \multirow{8}{*}{nuScenes}  

        & MapTRv2 \cite{liao2024maptrv2}$^\dagger$ & 24 & 0 & 26.8 & 14.7 & 44.4 & 28.6 & - & -\\
        & MapUnveiler \cite{kim2025unveiling}$^*$ & 24 & 3 & 43.2 & 26.5 & 48.7 & 39.4 & - & -\\
        \cline{2-10}
        & StreamMapNet \cite{yuan2024streammapnet}$^\dagger$ & 24 & 1 & 34.3 & 29.4 & 42.0 & 35.2 & 25.7 & 14.5 \\
        & \cellcolor{gray!20}StreamMapNet + MapTCL & \cellcolor{gray!20}24 & \cellcolor{gray!20}1 & \cellcolor{gray!20}39.2 & \cellcolor{gray!20}32.5 & \cellcolor{gray!20}45.0 & \cellcolor{gray!20}38.9\textcolor{ForestGreen}{(+3.7)} & \cellcolor{gray!20}28.5\textcolor{ForestGreen}{(+2.8)} & \cellcolor{gray!20}14.5 \\
        \cline{2-10}
        & SQD-MapNet \cite{wang2024stream}$^\dagger$ & 24 & 1 & 37.1 & 28.6 & 44.1 & 36.6 & 25.4 & 14.5 \\
        & \cellcolor{gray!20}SQD-MapNet + MapTCL & \cellcolor{gray!20}24 & \cellcolor{gray!20}1 & \cellcolor{gray!20}37.1 & \cellcolor{gray!20}32.0 & \cellcolor{gray!20}44.5 & \cellcolor{gray!20}37.9\textcolor{ForestGreen}{(+1.3)} & \cellcolor{gray!20}26.8\textcolor{ForestGreen}{(+1.4)} & \cellcolor{gray!20}14.5 \\
        \cline{2-10}
        & MapTracker \cite{chen2024maptracker}$^\dagger$ & 24 & 1 & 42.0 & 29.2 & 46.8 & 39.4 & 30.8 & 11.2 \\
        & \cellcolor{gray!20}MapTracker + MapTCL & \cellcolor{gray!20}24 & \cellcolor{gray!20}1 & \cellcolor{gray!20}44.2 & \cellcolor{gray!20}29.9 & \cellcolor{gray!20}48.9 & \cellcolor{gray!20}41.0\textcolor{ForestGreen}{(+1.6)} & \cellcolor{gray!20}32.4\textcolor{ForestGreen}{(+1.6)} & \cellcolor{gray!20}11.2 \\
        
        \midrule
        \midrule
        \multirow{7}{*}{Argoverse2}

        & MapTRv2 \cite{liao2024maptrv2}$^\dagger$ & 6 & 0 & 56.0 & 60.7 & 60.0 & 58.9 & - & - \\
        \cline{2-10}
        & StreamMapNet \cite{yuan2024streammapnet}$^\dagger$ & 6 & 1 & 49.1 & 51.9 & 59.7 & 53.6 & 35.4 & 14.9 \\
        & \cellcolor{gray!20}StreamMapNet + MapTCL & \cellcolor{gray!20}6 & \cellcolor{gray!20}1 & \cellcolor{gray!20}52.9 & \cellcolor{gray!20}55.1 & \cellcolor{gray!20}62.1 & \cellcolor{gray!20}56.7\textcolor{ForestGreen}{(+3.1)} & \cellcolor{gray!20}37.9\textcolor{ForestGreen}{(+2.5)} & \cellcolor{gray!20}14.9 \\
        \cline{2-10}
        & SQD-MapNet \cite{wang2024stream}$^\dagger$ & 6 & 1 & 52.4 & 53.0 & 60.8 & 55.4 &35.1 & 14.9 \\
        & \cellcolor{gray!20}SQD-MapNet + MapTCL & \cellcolor{gray!20}6 & \cellcolor{gray!20}1 & \cellcolor{gray!20}54.4 & \cellcolor{gray!20}55.7 & \cellcolor{gray!20}62.2 & \cellcolor{gray!20}57.4\textcolor{ForestGreen}{(+2.0)} & \cellcolor{gray!20}39.0\textcolor{ForestGreen}{(+3.9)} & \cellcolor{gray!20}14.9 \\
        \cline{2-10}
        & MapTracker \cite{chen2024maptracker}$^\dagger$ & 6 & 1 & 61.9 & 72.2 & 63.4 & 66.1 & 53.5 & 12.3 \\
        & \cellcolor{gray!20}MapTracker + MapTCL & \cellcolor{gray!20}6 & \cellcolor{gray!20}1 & \cellcolor{gray!20}62.5 & \cellcolor{gray!20}73.4 & \cellcolor{gray!20}64.7 & \cellcolor{gray!20}66.9\textcolor{ForestGreen}{(+0.8)} & \cellcolor{gray!20}54.6\textcolor{ForestGreen}{(+1.1)} & \cellcolor{gray!20}12.3 \\

        \bottomrule[1pt]
    \end{tabular}
    }
\vspace{-10pt}
\end{center}
\end{table*}

\begin{table*}[t]
\caption{Comparisons on nuScenes and Argoverse2 validation sets using geo-overlapping data split within $60m\times30m$ range. FPS is measured using a single NVIDIA RTX 3090 GPU. $\dagger$ results reproduced in our environment using available code, and $^*$ results from the corresponding papers. - indicates unavailable results (C-mAP not reported and FPS omitted due to different evaluation environments).}
\begin{center}
\label{tab:3_Comparison_oldsplit}
    \resizebox{0.75\textwidth}{!}
    {
    \setlength{\tabcolsep}{6pt}
    \renewcommand{\arraystretch}{1.3}
    \begin{tabular}{l|l|cc|cccc|c|c}
        \toprule[1pt]
        Dataset & Method & Epochs & \makecell{Fused \\ Frames} & AP$_{\textit{ped}}$ & AP$_{\textit{div.}}$ & AP$_{\textit{bound.}}$ & mAP & C-mAP & FPS \\
        \hline
        \multirow{9}{*}{nuScenes} 

        & MapTRv2 \cite{liao2024maptrv2}$^*$ & 24 & 0 & 59.8 & 62.4 & 62.4 & 61.5 & - & - \\
        & DTCLMapper \cite{li2024dtclmapper}$^*$ & 24 & 0 & 62.6 & 59.7 & 63.4 & 61.9 & -  & - \\
        & MapQR \cite{liu2024mapqr}$^*$ & 24 & 0 & 68.0 & 63.4 & 67.7 & 66.4 & -  & - \\
        & SQD-MapNet \cite{wang2024stream}$^\dagger$ & 24 & 1 & 59.7 & 62.7 & 61.9 & 61.4 & 44.5 & 14.5 \\
        
        & MapUnveiler \cite{kim2025unveiling}$^*$ & 24 & 3 & 67.6 & 67.6 & 68.8 & 68.0 & -  & - \\

        \cline{2-10}
        
        & StreamMapNet \cite{yuan2024streammapnet}$^\dagger$ & 24 & 1 & 58.9 & 60.2 & 58.7 & 59.3 & 44.9 & 14.5 \\
        
        & \cellcolor{gray!20}StreamMapNet + MapTCL & \cellcolor{gray!20}24 & \cellcolor{gray!20}1 & \cellcolor{gray!20}62.1 & \cellcolor{gray!20}62.9 & \cellcolor{gray!20}61.4 & \cellcolor{gray!20}62.1\textcolor{ForestGreen}{(+2.8)} & \cellcolor{gray!20}45.4\textcolor{ForestGreen}{(+0.5)} & \cellcolor{gray!20}14.5 \\
        \cline{2-10}
    
        & MapTracker \cite{chen2024maptracker}$^\dagger$ & 24 & 1 & 70.3 & 67.3 & 66.2 & 67.9 & 53.6 & 11.2 \\
         & \cellcolor{gray!20}MapTracker + MapTCL & \cellcolor{gray!20}24 & \cellcolor{gray!20}1 & \cellcolor{gray!20}72.3 & \cellcolor{gray!20}69.6 & \cellcolor{gray!20}68.8 & \cellcolor{gray!20}70.2\textcolor{ForestGreen}{(+2.3)} & \cellcolor{gray!20}59.8\textcolor{ForestGreen}{(+6.2)} & \cellcolor{gray!20}11.2 \\
        
        \midrule \midrule
        \multirow{9}{*}{Argoverse2} 
        
        & MapTRv2 \cite{liao2024maptrv2}$^*$ & 6 & 0 & 60.7 & 68.9 & 64.5 & 64.7 & - & - \\
        & DTCLMapper \cite{li2024dtclmapper}$^*$ & 6 & 0 & 69.4 & 61.9 & 64.1 & 65.1 & - & - \\
        & MapQR \cite{liu2024mapqr}$^*$ & 6 & 0 & 72.3 & 64.3 & 68.1 & 68.2 & -  & - \\
        & SQD-MapNet \cite{wang2024stream}$^\dagger$ & 6 & 1 & 54.0 & 52.3 & 59.3 & 55.2 & 35.0 & 14.9 \\
        
        & MapUnveiler \cite{kim2025unveiling}$^*$ & 6 & 3 & 68.9 & 73.7 & 68.9 & 70.5 & - & - \\
        \cline{2-10}
        & StreamMapNet \cite{yuan2024streammapnet}$^\dagger$ & 6 & 1 & 50.8 & 52.8 & 57.4 & 53.7 & 35.2 & 14.9 \\
        & \cellcolor{gray!20}StreamMapNet + MapTCL & \cellcolor{gray!20}6 & \cellcolor{gray!20}1 & \cellcolor{gray!20}54.0 & \cellcolor{gray!20}53.4 & \cellcolor{gray!20}59.1 & \cellcolor{gray!20}55.6\textcolor{ForestGreen}{(+2.9)} & \cellcolor{gray!20}37.7\textcolor{ForestGreen}{(+2.5)} & \cellcolor{gray!20}14.9 \\
        \cline{2-10}
        & MapTracker \cite{chen2024maptracker}$^\dagger$ & 6 & 1 & 66.3 & 72.9 & 64.5 & 68.1 & 53.5 & 12.3 \\
        & \cellcolor{gray!20}MapTracker + MapTCL & \cellcolor{gray!20}6 & \cellcolor{gray!20}1 & \cellcolor{gray!20}66.3 & \cellcolor{gray!20}73.6 & \cellcolor{gray!20}66.4 & \cellcolor{gray!20}68.7\textcolor{ForestGreen}{(+0.6)} & \cellcolor{gray!20}56.3\textcolor{ForestGreen}{(+2.8)} & \cellcolor{gray!20}12.3 \\
        
        \bottomrule[1pt]
    \end{tabular}
    }
\vspace{-15pt}
\end{center}
\end{table*}

\subsection{Experimental Settings}

\textbf{Dataset.} 
We conduct experiments on two benchmarks: nuScenes~\cite{caesar2020nuscenes} and Argoverse2~\cite{wilson2023argoverse}. nuScenes contains 1,000 driving scenes of about 20 seconds with 2 Hz annotations from six surround cameras at 480$\times$800 resolution. Argoverse2 includes 1,000 scenarios of about 15 seconds with 10 Hz annotations from seven surround cameras at 608$\times$608 resolution. Both datasets follow the standard 700/150/150 train/val/test split.
Following~\cite{yuan2024streammapnet}, the official splits of nuScenes and Argoverse2 have geographically overlapping areas, which can lead to model overfitting. Therefore, we mainly conduct training and evaluation on the newly designated non-overlapping (i.e., newsplit) datasets rather than oldsplit.

\textbf{Metrics.} Following prior works \cite{yuan2024streammapnet, liu2023vectormapnet, liao2022maptr, liao2024maptrv2, li2022hdmapnet}, we adopt Chamfer distance-based Average Precision (AP) as the primary evaluation metric. The AP is calculated at three distance thresholds: \{0.5m, 1.0m, 1.5m\}, and the final mean Average Precision (mAP) is derived by averaging the results across three road element categories: pedestrian crossings, lane dividers, and road boundaries. 
However, the Chamfer distance-based mAP evaluates only single-frame detection performance and does not consider temporal consistency in detecting and tracking the same map elements across frames. To address this limitation, MapTracker~\cite{chen2024maptracker} introduced a consistency-aware metric, termed C-mAP, which we adopt in this work.

\textbf{Implementation Details.}
We integrate MapTCL into various baseline models~\cite{yuan2024streammapnet, wang2024stream, chen2024maptracker} and compare its performance with other SOTA methods.
To ensure a fair comparison, we fix the fusion length to a single frame (Fused Frames = 1) for all temporally-aware baselines (StreamMapNet, SQD-MapNet, MapTracker) to demonstrate that MapTCL enhances temporal consistency without relying on extended temporal fusion.
Another temporal model, MapUnveiler, is excluded from this setting due to its unavailable code.

StreamMapNet and SQD-MapNet are trained with a two-stage schedule (single-frame followed by temporal training), and MapTCL is applied only in the temporal stage. They are trained for 24 epochs on nuScenes (6/18) and 6 epochs on Argoverse2 (1/5). MapTracker uses a three-stage schedule (6/2/16 on nuScenes and 2/1/3 on Argoverse2), with MapTCL applied only in the final stage. This design is adopted to reduce the risk of error propagation, since early training stages typically contain many false positive predictions that can lead to unstable temporal associations.


All models are trained on 4 NVIDIA A6000 GPUs with a batch size of 16 using the AdamW optimizer~\cite{loshchilov2017fixing}. We set the number of predicted instances to $N=100$ with each instance represented by $N_p=20$ vector points. We also set $N_t=5$, $\tau=0.3$, $\lambda_1=50.0$, $\lambda_2=5.0$, $\lambda_3=0.1$ and $\beta=1.0$ in StreamMapNet + MapTCL.
Other optimizer settings and hyperparameters vary across experiments. For example, on Argoverse2 newsplit with StreamMapNet + MapTCL, we use an initial learning rate of $5e-4$ with a weight decay of 0.03. We also set $w_{1:5} = [9, 7, 5, 3, 1]$ and $u_{1:5} = a_{1:5} = [1.0, 0.7, 0.5, 0.3, 0.1]$.
Regarding ego-motion-based warping, MapTCL aligns past information to the current frame for temporal consistency. StreamMapNet and SQD-MapNet assume accurate ego poses, while MapTracker introduces pose noise during training to improve robustness to pose estimation errors.


 

\vspace{-4pt}
\subsection{Main Results}
\vspace{-1pt}
MapTCL consistently improves the performance of diverse baseline models across two benchmark datasets under both newsplit and oldsplit settings.

\begin{figure*}[t]
\begin{center}

\includegraphics[width=0.80\linewidth]{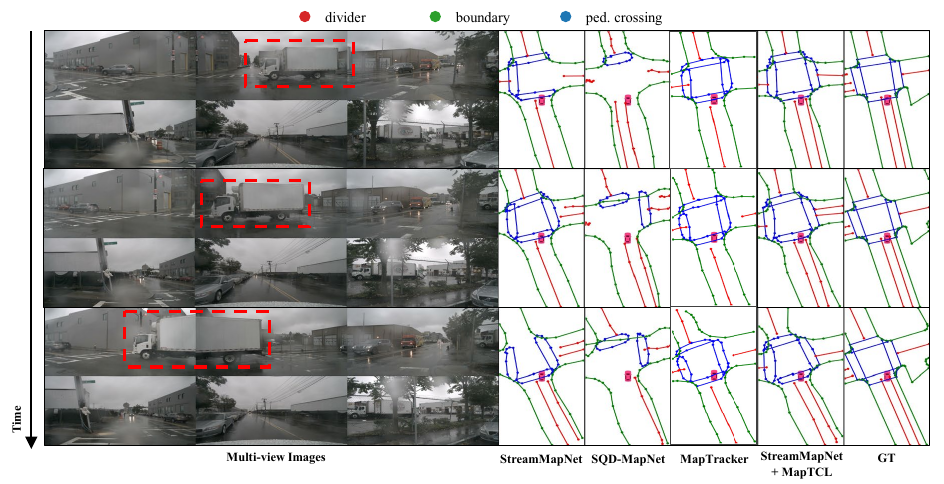}
\vspace{-10pt}
\caption{Qualitative comparisons on newsplit of nuScenes validation within $60m\times30m$ perception range. StreamMapNet + MapTCL is compared with other baselines (StreamMapNet \cite{yuan2024streammapnet}, SQD-MapNet \cite{wang2024stream}, MapTracker \cite{chen2024maptracker}) and ground truth. The red dashed boxes indicate occlusions caused by surrounding objects, which lead to temporal inconsistency.
}

\label{fig:Qualitative_nuscenes}
\end{center}
\vspace{-18pt}
\end{figure*}

\begin{figure}[t]
\begin{center}
\includegraphics[width=0.93\linewidth]{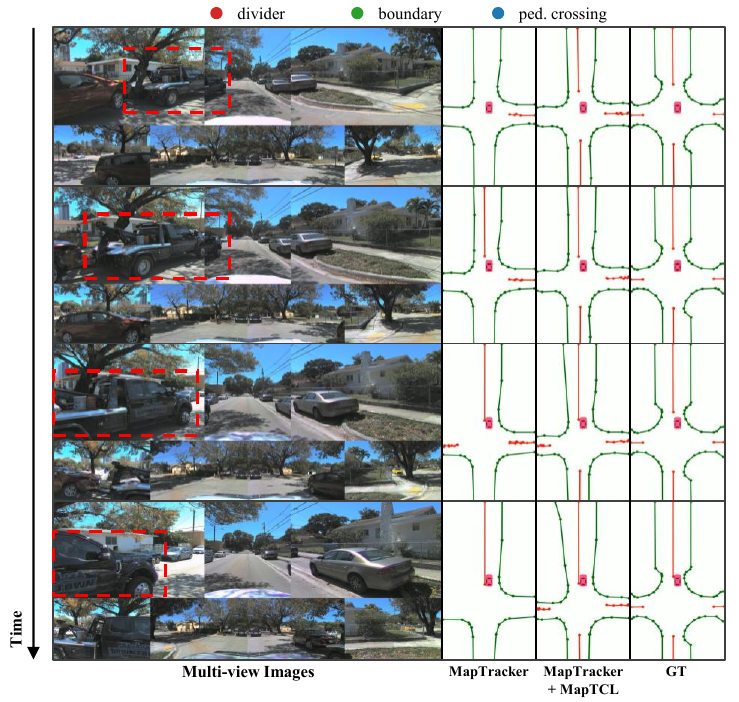}
\vspace{-9pt}
\caption{Qualitative comparisons on newsplit of Argoverse2 validation within $60m\times30m$ perception range. MapTracker + MapTCL is compared with the baseline and ground truth. The red dashed boxes indicate occlusions caused by surrounding objects, which lead to temporal inconsistency.}
\label{fig:Qualitative_argo}
\end{center}
\vspace{-12pt}
\end{figure}

\textbf{Results on Non-overlapping Split.}
Table~\ref{tab:3_Comparison_newsplit} presents the quantitative results on the nuScenes and Argoverse2 newsplit validation sets.
On nuScenes, MapTCL improves the mAP of StreamMapNet from 35.2 to 38.9 (+3.7) and SQD-MapNet from 36.6 to 37.9 (+1.3). Similarly, MapTracker achieves an improvement from 39.4 to 41.0 (+1.6). In addition to accuracy gains, temporal consistency is also enhanced, with C-mAP increasing by +2.8, +1.4, and +1.6 for StreamMapNet, SQD-MapNet, and MapTracker, respectively. Similar improvements are also observed on Argoverse2, where MapTCL consistently enhances both mAP and C-mAP across all baselines.
Notably, MapTracker + MapTCL surpasses MapUnveiler~\cite{kim2025unveiling}, a SOTA model utilizing three fused frames, on nuScenes despite operating under a single-frame fusion setting. These improvements are achieved without any additional inference costs, as indicated by the identical FPS after applying MapTCL. 
Fig.~\ref{fig:Qualitative_nuscenes} shows qualitative comparisons between StreamMapNet + MapTCL, baseline methods (StreamMapNet, SQD-MapNet, and MapTracker), and the ground truth on the nuScenes newsplit dataset. Among the $N$ predicted vector instances, only those with confidence scores above 0.4 are visualized. As shown in the figure, our method generates more accurate and temporally consistent HD maps under challenging conditions such as dynamic occlusions and rainy weather. In contrast, other methods exhibit temporally inconsistent predictions, where certain map elements disappear, deform, or become misaligned across frames.
Fig.~\ref{fig:Qualitative_argo} further presents qualitative comparisons between MapTracker and MapTracker + MapTCL on the Argoverse2 dataset. Similarly, MapTCL demonstrates improved temporal consistency even in the presence of occlusions caused by surrounding dynamic objects.

\begin{table}[t]
\caption{Performance on newsplit of nuScenes and Argoverse2 within $100m\times50m$ perception range.}
\vspace{-8pt}
\label{tab:abl_long_range}
\begin{center}
\resizebox{0.85\columnwidth}{!}{
\renewcommand{\arraystretch}{1.0}
\begin{tabular}{l|l|cccc}
    \toprule[1pt]
    Dataset & Method & AP$_{\textit{ped}}$ & AP$_{\textit{div.}}$ & AP$_{\textit{bound.}}$ & mAP \\
    \hline
    \multirow{2}{*}{nuScenes} & 
    StreamMapNet & 27.1 & 18.5 & 21.5 & 22.4 \\
    & \cellcolor{gray!20}\textbf{+ MapTCL}  & \cellcolor{gray!20}\textbf{29.4} & \cellcolor{gray!20}\textbf{21.8} & \cellcolor{gray!20}\textbf{28.3} & \cellcolor{gray!20}\textbf{26.5} \\
    \midrule \midrule
    \multirow{2}{*}{Argoverse2} & 
    StreamMapNet & 53.5 & 42.2 & 46.7 & 47.5 \\
    & \cellcolor{gray!20}\textbf{+ MapTCL} & \cellcolor{gray!20}\textbf{56.9} & \cellcolor{gray!20}\textbf{44.2} & \cellcolor{gray!20}\textbf{50.0} & \cellcolor{gray!20}\textbf{50.2} \\
    \bottomrule[1pt]
\end{tabular}
}
\vspace{-25pt}
\end{center}
\end{table}

\begin{table}[t]
\begin{center}
\caption{Performance under heavy occlusions on newsplit of nuScenes within $60m\times30m$ perception range.}
\label{tab:abl_occlusion}    
\resizebox{0.95\columnwidth}{!}{ 
\begin{tabular}{l|cccc|c}
        \toprule
        \makecell{Method} &
        AP$_{\textit{ped}}$ & 
        AP$_{\textit{div.}}$ & 
        AP$_{\textit{bound.}}$ & 
        mAP & C-mAP   \\
        \midrule
         StreamMapNet & 29.7 & 23.4 & 35.1 & 29.5 & 29.9\\
         SQD-MapNet & 28.5 & 24.5 & 36.1 & 29.7 & 29.0\\
         \rowcolor{gray!20} StreamMapNet + MapTCL & \textbf{33.3} & \textbf{27.2} & \textbf{36.7} & \textbf{32.4} & \textbf{32.8}\\
        \bottomrule
\end{tabular}
}
\vspace{-20pt}
\end{center}
\end{table}

\begin{table}[t]
\caption{Component ablation of our method on newsplit of nuScenes within $60m\times30m$ perception range.}
\vspace{-14pt}
\label{tab:abl_component}
\begin{center}
\resizebox{0.83\columnwidth}{!}{
\renewcommand{\arraystretch}{1.0} 
\begin{tabular}{cccc|cccc}
        \toprule[1pt]
         \makecell{VCL} & 
         \makecell{BVCL} & 
         \makecell{SL} & 
         \makecell{RCL} & AP$_{\textit{ped}}$ & AP$_{\textit{div.}}$ & AP$_{\textit{bound.}}$ & mAP    \\\hline
         & & & & 32.7 & 29.7 & 41.9 & 35.2 \\
         \cmark &  &  &  & 37.2 & 30.2 & 40.2 & 35.9 \\
          & \cmark &  &  & 38.6 & 30.2 & 43.0 & 37.2 \\
          &  & \cmark &  & 38.7 & 31.5 & 43.4 & 37.8\\ 
          &  & \cmark & \cmark & 37.2 & 31.9 & \textbf{45.8} & 38.2\\ 
          \rowcolor{gray!20} & \cmark & \cmark & \cmark & \textbf{39.2} & \textbf{32.5} & 45.0 & \textbf{38.9}\\ 
        \bottomrule[1pt]
    \end{tabular}
}
\vspace{-10pt}
\end{center}
\end{table}

\vspace{-2pt}
\textbf{Results on Original Split.}
As shown in Table~\ref{tab:3_Comparison_oldsplit}, MapTCL consistently improves both mAP and C-mAP across all baselines on the original splits of nuScenes and Argoverse2.
In particular, MapTracker + MapTCL achieves the largest C-mAP gain on nuScenes. 
It may benefit more from the geographically overlapping split, where train-validation overlap can make temporal correspondences easier to exploit. This large gain may also be affected by hyperparameter settings, such as confidence threshold and loss weights.

\textbf{Performance on Long Perception Range.}
Table~\ref{tab:abl_long_range} shows that our method consistently outperforms the baseline under a $100m\times50m$ perception range, demonstrating its effectiveness in long-range perception scenarios.

\textbf{Performance on Occlusion Scenario} 
Table~\ref{tab:abl_occlusion} reports the performance on the nuScenes validation set under occlusion-prone scenarios, where we select scenes containing dynamic objects within $5m$ of the ego vehicle. Our method consistently outperforms both StreamMapNet and SQD-MapNet,
demonstrating its robustness in occluded environments.
\vspace{-4pt}
\subsection{Ablation Studies}
\vspace{-3pt}
\textbf{Component Ablation.} 
\begin{table}[t]
\caption{Memory size ablation of our method on newsplit of Argoverse2 within $60m\times30m$ perception range.}
\vspace{-14pt}
\label{tab:abl_memory_size}
\begin{center}
    \resizebox{0.80\columnwidth}{!}{  
    \renewcommand{\arraystretch}{1.0}  
    \begin{tabular}{c|cccc|c}
        \toprule[1pt]
         \makecell{$N_t$} & 
         AP$_{\textit{ped}}$ & AP$_{\textit{div.}}$ & AP$_{\textit{bound.}}$ & mAP & C-mAP    \\\hline
         1 & 52.3 & 53.8 & 61.7 & 55.9 & 36.8\\
         3 & 51.5 & 53.9 & \textbf{62.5} & 56.0 & 36.5\\
         \rowcolor{gray!20} 5 & \textbf{52.9} & \textbf{55.1} & 62.1 & \textbf{56.7} & \textbf{37.9}\\ 
         7 & 45.8 & 50.5 & 58.6 & 51.6 & 32.5\\
        \bottomrule[1pt]
    \end{tabular}
    }
\vspace{-18pt}
\end{center}
\end{table}
We analyze the contribution of each component, including Vector Consistency Learning (VCL), Bidirectional Vector Consistency Learning (BVCL), Segmentation Learning (SL), and Raster Consistency Learning (RCL), as summarized in Table~\ref{tab:abl_component}.
VCL is a simplified BVCL variant that only performs forward matching, omitting backward matching.
SL refers to performing only segmentation learning within RCL without temporal learning. 
All components improve performance, with the BVCL and RCL combination yielding the highest gain of 3.7 mAP.

\vspace{-1pt}
\textbf{Memory Ablation.}
Table~\ref{tab:abl_memory_size} shows the effect of memory size used in MapTCL for computing the temporal consistency loss. Increasing the memory size improves performance by incorporating longer temporal context, with the best results achieved at a memory size of 5. However, using too many past frames may introduce outdated or noisy information, which can degrade performance.
\begin{table}[t]
\caption{Ablation study on confidence threshold of BVCL on newsplit of Argoverse2.}
\vspace{-12pt}
\label{tab:abl_confidence}
\begin{center}
    
    \resizebox{0.80\columnwidth}{!}{  
    \renewcommand{\arraystretch}{1.0}  
    \begin{tabular}{c|cccc|c}
        \toprule[1pt]
         \makecell{$\tau$} & 
         AP$_{\textit{ped}}$ & AP$_{\textit{div.}}$ & AP$_{\textit{bound.}}$ & mAP & C-mAP    \\\hline
         0.1 & 49.7 & 52.3 & 59.9 & 54.0 & 23.8 \\
         \rowcolor{gray!20} 0.3 & \textbf{52.9} & \textbf{55.1} & 62.1 & \textbf{56.7} & \textbf{37.9}\\ 
         0.5 & 51.5 & 53.9 & \textbf{62.5} & 56.0 & 36.5\\
        \bottomrule[1pt]
    \end{tabular}
    }
\end{center}
\vspace{-20pt}
\end{table}

\vspace{-1pt}
\textbf{Confidence Threshold Ablation in BVCL.}
We further evaluate the effect of different confidence thresholds in BVCL. Only instances with confidence scores above the threshold are used to compute the consistency loss. As shown in Table~\ref{tab:abl_confidence}, both overly low and overly high thresholds degrade performance, with the best results achieved at a threshold of 0.3.


\vspace{-7pt}
\section{CONCLUSION}
\label{sec:conclusion}

We propose MapTCL, an auxiliary training strategy that enhances temporal consistency in online HD maps by modeling consistency losses across past frames. By incorporating BVCL and RCL, MapTCL improves geometric and semantic coherence between consecutive predictions and consistently reduces temporal jitter, without introducing additional inference cost. Experiments on nuScenes and Argoverse2 show consistent improvements in both accuracy and temporal stability. However, MapTCL is sensitive to hyperparameter settings, especially the temporal context length, where overly long histories may introduce noisy supervision and degrade performance.

\addtolength{\textheight}{-12cm}   




\bibliographystyle{ieeetr} 
\bibliography{ref}

\end{document}